\documentclass{article}

\usepackage{microtype}
\usepackage{graphicx}
\usepackage{subcaption}
\usepackage[T1]{fontenc}
\usepackage{booktabs} 
\usepackage{amsfonts,psfrag,epsfig}
\usepackage{color}
\usepackage{amsmath,amssymb,amsthm}
\usepackage{tabularx}
\usepackage{hyperref}

\newtheorem{theorem}{Theorem}
\newtheorem{claim}[theorem]{Claim}

\usepackage{mathtools}

\DeclarePairedDelimiterX{\inp}[2]{\langle}{\rangle}{#1, #2}

\usepackage[accepted]{icml2019}

\icmltitlerunning{Spectral Approximate Inference} 

\begin{document}

\twocolumn[
\icmltitle{Spectral Approximate Inference} 



\icmlsetsymbol{equal}{*}

\begin{icmlauthorlist}
\icmlauthor{Sejun Park}{ee}
\icmlauthor{Eunho Yang}{cs,ai,aitrics}
\icmlauthor{Se-Young Yun}{ai,ie}
\icmlauthor{Jinwoo Shin}{ee,ai,aitrics}
\end{icmlauthorlist}
\icmlaffiliation{ee}{School of Electrical Engineering, KAIST, Daejeon, Korea}

\icmlaffiliation{cs}{School of Computing, KAIST, Daejeon, Korea}

\icmlaffiliation{ie}{Department of Industrial \& System Engineering, KAIST, Daejeon, Korea}

\icmlaffiliation{ai}{Graduate School of AI, KAIST, Daejeon, Korea}

\icmlaffiliation{aitrics}{AITRICS, Seoul, Korea}

\icmlcorrespondingauthor{Jinwoo Shin}{jinwoos@kaist.ac.kr}

\icmlkeywords{Machine Learning, ICML}

\vskip 0.3in
]



\printAffiliationsAndNotice{}  

\begin{abstract}
Given a graphical model (GM),
computing its partition function is the most essential inference task,
but it is computationally intractable in general.
To address the issue, iterative approximation algorithms exploring certain local structure/consistency of GM
have been investigated as popular choices in practice.
However, due to their local/iterative nature, they often output poor approximations or even do not converge, e.g.,
in low-temperature regimes (hard instances of large parameters).
To overcome the limitation, we propose a novel approach utilizing the global spectral feature of GM.
Our contribution is two-fold:
(a) we first propose a fully polynomial-time approximation scheme (FPTAS) for approximating the partition function of GM 
associating with a low-rank coupling matrix;
(b) for general high-rank GMs,
we design a spectral mean-field scheme utilizing (a) as a subroutine, where
it approximates a high-rank GM into a product of rank-1 GMs for an efficient approximation of the partition function.
The proposed algorithm is more robust in its running time and accuracy than prior methods, i.e., 
neither suffers from the convergence issue nor depends on hard local structures, as demonstrated in
our experiments.
\end{abstract}

\section{Introduction}
Graphical models (GMs) provide a succinct representation of a joint probability distribution over a set of random variables by encoding their conditional dependencies in graphical structures.
GMs have been studied in various fields of machine learning, including computer vision \cite{freeman2000learning}, speech recognition \cite{bilmes2004graphical}
and deep learning \cite{salakhutdinov2010efficient}.
Most inference problems arising in GMs, e.g., obtaining desired samples and computing marginal distributions,
can be easily reduced to computing their partition function (normalizing constant).
However, computing the partition function is \#P-hard in general even to approximate \cite{jerrum1993polynomial}, 
which is thus a fundamental barrier for inference tasks of GM.

Variational inference is one of the most popular heuristics in practice
for estimating the partition function.
It is typically achieved via running iterative local message-passing algorithms, e.g.,
mean-field approximation \cite{parisi1988statistical,jain2018mean} and belief propagation \cite{pearl1982reverend,wainwright2005new}. 
Markov chain Monte Carlo (MCMC) method \cite{neal2001annealed,efthymiou2016convergence} 
is another popular approach,
where 
it usually samples from GMs via Markov chains with a local transition, e.g., Gibbs sampler \cite{geman1984stochastic}, and estimates 
a target expectation by averaging over samples.
Unfortunately, both variational and MCMC methods are hard to guarantee the convergence/mixing  under some fixed computation budget and known to
output poor approximation in the low-temperature regime, i.e., large parameters of GM, due to the non-existence of
the so-called correlation decay \cite{weitz2006counting,bandyopadhyay2008counting}.
On the other hand, variable elimination \cite{dechter1999bucket,dechter2003mini,liu2011bounding,xue2016variable,wrigley2017tensor,ahn2018gauged,ahn2018bucket} is one of popular `convergence free' methods for approximating the partition function. At each step, it sequentially marginalizes a chosen variable and generates complex high-order factors approximating the marginalized variable and its associated factors.
Hence, it guarantees to terminate after marginalizing all variables.
However,
the performance of variable elimination schemes is also significantly degraded in the low-temperature regime, due to its local/iterative nature of processing
variables one by one.
\begin{figure*}[ht]
\centering
    \includegraphics[width=0.91\textwidth]{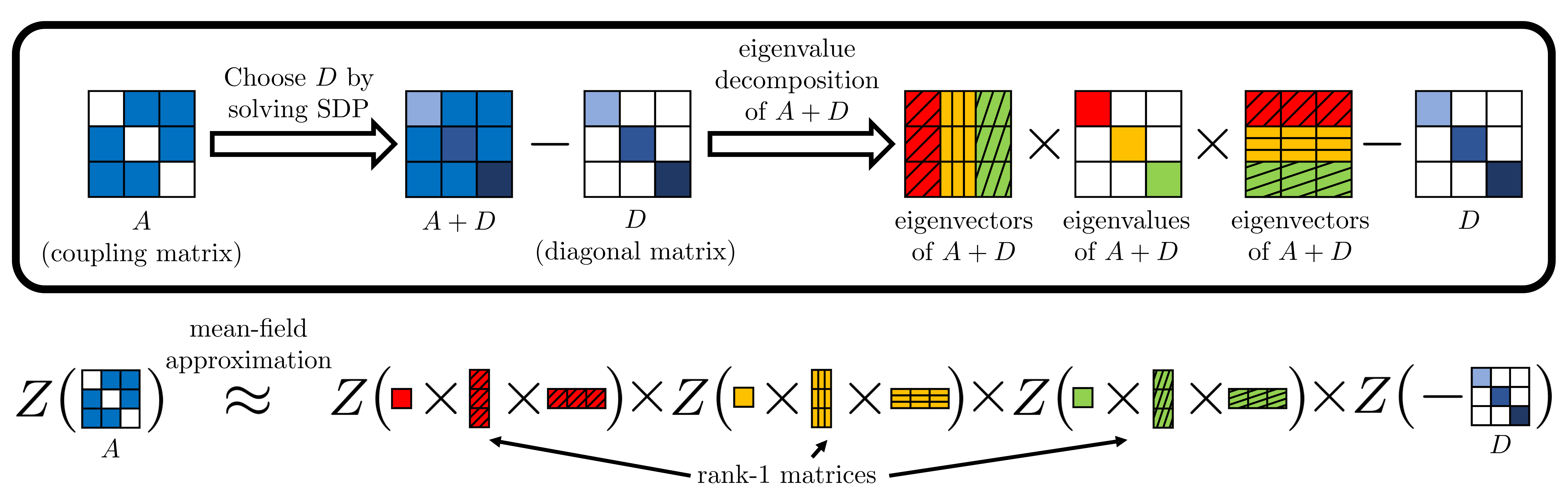}
    \caption{An illustration of the proposed partition function approximation scheme.}
    \label{fig:alg}
\end{figure*}

{\bf Contribution.} In this paper, we propose a completely new approach 
by investigating the global information of GM, to overcome the limitation of prior methods.
To this end, we study the spectral feature of the coupling matrix of GM and propose a partition function approximation algorithm utilizing the eigenvectors and eigenvalues.
In particular, 
if the matrix-rank and parameters of GM are bounded, i.e., $O(1)$, then 
we prove that the proposed algorithm is a fully polynomial-time approximation scheme (FPTAS), 
even for GMs with high treewidth.
Such polynomial-time approximation schemes have been typically investigated in the literature under
certain structured GMs \cite{temperley1961dimer, pearl1982reverend, dechter1999bucket,jerrum2004polynomial},
and
high-temperature regimes \cite{zhang2011approximating,li2013correlation,patel2017deterministic}  
or homogeneity of GM parameters \cite{jerrum1993polynomial,sinclair2014approximation,molkaraie2016importance,liu2017ising,patel2017deterministic,molkaraie2018monte}.
Our theoretical result provides a new class of GMs for the direction.

Despite the theoretical value of the proposed algorithm for low-rank GMs,
it is very expensive to run for
general high-rank GMs as its complexity grows exponentially with respect to the rank.
To address this issue, we decompose the partition function of high-rank GM into a product of those of rank-1 GMs. Then, we run the proposed FPTAS algorithm to compute all rank-1 partition functions and combine them to approximate the original partition function.
For improving our approximation, we additionally suggest running a semi-definite programming to discover a better
spectral decomposition of the partition function.
In a sense, our approach is of mean-field type, but different from the traditional ones decomposing GM itself without spectral pre-processing.
We present an illustration of the proposed scheme in Figure \ref{fig:alg}.

The proposed mean-field scheme can be universally applied to any GMs without the rank restriction.
Its computational complexity 
scales well for large GMs without suffering from the convergence issue.
Furthermore,
its approximation quality is quite robust against hard GM instances of heterogeneous parameters
since the utilized spectral feature grows linearly with respect to the inverse temperature, i.e., scale of parameters. 
Our experiments demonstrate that the proposed scheme indeed outperforms
mean-field approximation, belief propagation and variable elimination, in particular, significantly in the low-temperature regimes where the prior methods fail.

\section{Spectral Inference for Low-Rank GMs}\label{sec:lowrank}
We begin with introducing the definition of the pairwise binary graphical model (GM).
Given a vector $\boldsymbol\theta\in\mathbb{R}^n$ and a symmetric matrix $A\in\mathbb{R}^{n\times n}$, we define GM as the following joint distribution on $\mathbf{x}\in\Omega:=\{-1,1\}^n$:
\begin{align}
   \mathbb{P}(\mathbf x)=\frac1Z\exp\left(\inp{\boldsymbol\theta}{\mathbf x}+\mathbf{x}^TA\mathbf{x}\right)\label{eq:gm}
\end{align}
where $\inp{\cdot}{\cdot}$ denotes the inner product and $Z$ is the normalizing constant.
The above definition of GM coincides with the following conventional definition associating with an undirected graph $\mathcal G=(\mathcal V,\mathcal E)$ defined as:
\begin{align}
   \mathbb{P}(\mathbf x)\propto\exp\left(\sum_{i\in \mathcal V}\theta_ix_i+2\sum_{(i,j)\in \mathcal E}A_{ij}x_ix_j\right).\label{eq:gm2}
\end{align}
where $\mathcal V=\{1,\dots,n\}$ and $\mathcal E=\{(i,j):A_{ij}\ne0,~i<j\}$.

The normalizing constant $Z$ of \eqref{eq:gm} is called the partition function defined as follows:
\begin{equation}\label{eq:z}
    Z=Z(\boldsymbol\theta, A):=\sum_{\mathbf x\in\Omega}\exp\left(\inp{\boldsymbol\theta}{\mathbf x}+\mathbf{x}^TA\mathbf x\right).
\end{equation}
Computing $Z$ is one of the most essential inference tasks arising in GMs.
However, it is known to be computationally intractable in general, i.e., \#P-hard even to approximate \cite{jerrum1993polynomial}.
In particular, the case when the magnitudes of entries of $A$ are large is called, the low-temperature regime \cite{sykes1965derivation},
where $Z$ is known to be harder to approximate provably \cite{sly2012computational,galanis2014inapproximability}.
This is indeed the regime where known heuristics also fail badly.

In this section, we show that $Z$ is possible to be approximated in 
polynomial-time if there exists a diagonal matrix $D$ such that
the rank of $A+D$ is bounded, i.e., $O(1)$. 
Just for clarity, we primarily focus on the case when $A$ is of low-rank itself (i.e., $D=0$)
and then describe at the end of this section how our results are extended to the case when $A+D$ is of low-rank for any diagonal matrix $D$.

\subsection{Overall Approach: Approximate Inference via Spectral Decomposition}\label{sec:overspec}
To design such a polynomial-time algorithm,
we first reformulate $Z$ using the eigenvalues/eigenvectors of $A$ as follows:
\begin{align}
Z&=\sum_{\mathbf{x}\in\Omega}\exp\left(\inp{\boldsymbol\theta}{\mathbf{x}}+\mathbf{x}^TA\mathbf{x}\right)\notag\\
&=\sum_{\mathbf{x}\in\Omega}\exp\left(\inp{\boldsymbol\theta}{\mathbf{x}}+\sum_{j=1}^r\lambda_j\inp{\mathbf{v}_j}{\mathbf{x}}^2\right)\label{eq:z2}
\end{align}
where $\lambda_j$ and $\mathbf{v}_j$ denote the $j$-th largest non-zero eigenvalue and its corresponding unit eigenvector of $A$ and $r$ denotes the rank of $A$.
We note that such a decomposition is always possible because $A$ is a real symmetric matrix, i.e., all eigenvalues are real.
However, even with a small rank $r$, a naive computation of $Z$ is still intractable as it is a summation over exponentially many terms. Our main idea is approximating $\lambda_j\inp{\mathbf{v}_j}{\mathbf{x}}^2$ in \eqref{eq:z2} to its quantized value in order to drastically reduce the number of summations. Toward this, we rewrite \eqref{eq:z2} as
\begin{align*}
&Z=\sum_{\mathbf{x}\in\Omega}\exp\left(\inp{\boldsymbol\theta}{\mathbf{x}}+\sum_{j=1}^r\text{sign}(\lambda_j)\inp*{\mathbf{u}_j}{\mathbf{x}}^2\right)
\end{align*}
where $\text{sign}(\lambda_j)\in\{-1,1\}$ denotes the sign of $\lambda_j$ and $\mathbf{u}_j=\sqrt{|\lambda_j|}\mathbf{v}_j$. Here we deliberately choose some mapping $f_j:\Omega \rightarrow\mathbb{Z}$ (it will be explicitly described in Section \ref{sec:choicef}) so that $c\cdot f_j(\mathbf{x})\approx\inp{\mathbf{u}_j}{\mathbf{x}}$ for some fixed constant $c>0$ and hence $Z$ can be nicely approximated as
\begin{align}
\sum_{\mathbf{x}\in\Omega}\exp\big(\inp{\boldsymbol\theta}{\mathbf{x}}\big)\exp\left(\sum_{j=1}^r\text{sign}(\lambda_j)\big(c\cdot f_j(\mathbf{x})\big)^2\right).\label{eq:lowrankz-1}
\end{align}
Note that $c$ decides a quantization interval and $c\cdot f_j(\mathbf{x})$ represents a quantized value of $\inp{\mathbf{u}_j}{\mathbf{x}}$.
Namely, for each $\mathbf{x}\in\Omega$, we will design $\mathbf{f}(\mathbf{x})=[f_j(\mathbf{x})]_{j=1}^r\in\mathbb{Z}^r$ for approximating $\inp{\mathbf{u}_j}{\mathbf{x}}$ for all $j$.

Given such $\mathbf{f}$, we further process \eqref{eq:lowrankz-1} as
\begin{align}
&\sum_{\mathbf{x}\in\Omega}\exp\big(\inp{\boldsymbol\theta}{\mathbf{x}}\big)\exp\left(\sum_{j=1}^r\text{sign}(\lambda_j)\big(c\cdot f_j(\mathbf{x})\big)^2\right)\notag\\
&=\sum_{\mathbf{k}\in \mathbf{f}(\Omega)}\left(\sum_{\mathbf{x}\in \mathbf{f}^{-1}(\mathbf{k})}\exp\big({\inp{\boldsymbol\theta}{\mathbf{x}}}\big)\right)\notag\\
&\qquad\qquad\qquad\times\exp\left(\sum_{j=1}^r\text{sign}(\lambda_j)(c\cdot k_j)^2\right)\notag\\
&=\sum_{\mathbf{k}\in \mathbf{f}(\Omega)}t(\mathbf k)\exp\left(\sum_{j=1}^r\text{sign}(\lambda_j)(c\cdot k_j)^2\right).\label{eq:lowrankz-2}
\end{align}
In the above, the first equality is from
replacing the summation over $\Omega$ by that over $\mathbf f(\Omega)$, i.e.,
for 
$\mathbf k=[k_j]_{j=1}^r\in \mathbf f(\Omega)$, each $k_j$ represents a possible value of $f_j(\mathbf{x})$. 
For the second equality, we define 
$t(\mathbf k):=\sum_{\mathbf x\in \mathbf{f}^{-1}(\mathbf k)}\exp\big(\inp{\boldsymbol\theta}{\mathbf x}\big).$
Finally, from \eqref{eq:lowrankz-1} and \eqref{eq:lowrankz-2}, 
one can observe that if $t(\mathbf k)$ is easy to compute and the cardinality of $\mathbf f(\Omega)$ is small, then 
the partition function $Z$ can be efficiently approximated.
In the following section, we provide more details on how to choose $\mathbf f$ for the desired property.

\subsection{How to Choose $\mathbf f$ and Compute $t(\mathbf k)$}\label{sec:choicef}
{\bf Choice of $\mathbf f$.} A naive choice of $\mathbf f$ can be 
\begin{equation}\label{eq:approxf-1}
f_j(\mathbf x)=\arg\min_{k_j\in\mathbb{Z}}|c\cdot k_j-\inp{\mathbf{u}_j}{\mathbf x}|
\end{equation}
for all $j$.
However, with the above choice of $\mathbf f$, 
it is unclear how to compute 
$t(\mathbf k)$ efficiently (in polynomial-time).
To address the issue, we propose a recursive construction of $\mathbf f$
by relaxing \eqref{eq:approxf-1}:
we
iteratively define $\mathbf f(\mathbf x)$ for $\mathbf x\in \mathcal{S}_i\setminus \mathcal{S}_{i-1}$ where $\mathcal{S}_i:=\{\mathbf x\in\Omega:x_\ell=-1,~\forall \ell>i\}$
so that
$$\{(-1,\dots,-1)\}=\mathcal{S}_0\subset \mathcal{S}_1\subset\dots\subset \mathcal{S}_{n-1}\subset \mathcal{S}_n =\Omega.$$

First, we define $\mathbf f$ for $\mathcal{S}_0$ following \eqref{eq:approxf-1}:
\begin{equation}\label{eq:approxf-3}
f_j\big((-1,\dots,-1)\big):=\arg\min_{k_j\in\mathbb{Z}}\left|c\cdot k_j+\sum_{i=1}^nu_{ji}\right|,
\end{equation}
for all $j$.
The construction of $\mathbf f$ for the rest $\mathcal{S}_n\setminus \mathcal{S}_0$ will be done in a recursive manner.
Suppose that $\mathbf f(\mathbf x)$ is defined for $\mathbf x\in \mathcal{S}_{i-1}$.
Then, 
we define $\mathbf f$ for $\mathbf x\in \mathcal{S}_i\setminus \mathcal{S}_{i-1}$ as follow:
\begin{equation}\label{eq:approxf-2}
f_j(\mathbf x):=f_j(\mathbf x^\prime)+\widehat{u}_{ji}
\end{equation}
where we define $\widehat{u}_{ji}:=\arg\min_{\widehat{u}_{ji}\in\mathbb{Z}}|c\cdot\widehat{u}_{ji}-2u_{ji}|$ and $\mathbf{x}^\prime\in \mathcal{S}_{i-1}$ such that $x^\prime_\ell=x_\ell$ except for $\ell=i$, i.e., $x^\prime_i=-1$.
Here, \eqref{eq:approxf-2} is motivated by the following approximation:
$c\cdot f_j(\mathbf{x}^\prime)\approx\inp{\mathbf{u}_j}{\mathbf{x}^\prime}$ and the definition of $\widehat{u}_{ji}$ implies that
\begin{align*}
c\cdot (f_j(\mathbf{x}^\prime)+\widehat{u}_{ji})\approx\inp{\mathbf{u}_j}{\mathbf{x}^\prime}+2u_{ji}=\inp{\mathbf{u}_j}{\mathbf x}
\end{align*}
where the equality is due to $x_i=1$ and $x^\prime_i=-1$.

In essence, we have so far constructed $\mathbf f$ via a dynamic programming to approximate  \eqref{eq:approxf-1},
which allows us to compute $t(\mathbf k)$ efficiently.
Furthermore, our choice of $\mathbf f$ ensures that $|\mathbf f(\Omega)|$ is bounded.
{Before describing how to compute $t(\mathbf k)$, let us discuss the bound of $|\mathbf f(\Omega)|$.
For bounding $|\mathbf f(\Omega)|$, we discover a bounded set $\mathcal B\subset\mathbb{Z}^r$ so that $\mathbf f(\Omega)\subset\mathcal B$ instead of characterizing $|\mathbf f(\Omega)|$ directly.
We explicitly describe such $\mathcal B$ as follows. }
\begin{claim}\label{claim:b}
$\mathbf f(\Omega)\subset\mathcal B$ where
\begin{align*}
\mathcal B&:=\prod_{j=1}^r\{-b_j,-b_j+1,\dots,b_j -1,b_j\},\\
b_j&:=\big\lceil\|\mathbf{u}_j\|_1/c+(n+1)/2\big\rceil.
\end{align*}
Furthermore, $|\mathcal B|$ is bounded by
\begin{align*}
|\mathcal B|\le2^r\prod_{j=1}^r\left(\frac1c\sqrt{|\lambda_j|n}+\frac{n}2+1\right).
\end{align*}
\end{claim}
We present the proof of Claim \ref{claim:b} in the supplementary material.
Finally, given $t(\mathbf k)$ and $\mathcal B$ as defined in Claim \ref{claim:b}, we approximate the partition function $Z$ as follows (see \eqref{eq:lowrankz-1} and \eqref{eq:lowrankz-2}):
\begin{equation*}
Z\approx\sum_{\mathbf k\in\mathcal B}t(\mathbf k)\exp\left(\sum_{j=1}^r\text{sign}(\lambda_j)(c\cdot k_j)^2\right),
\end{equation*}
where $t(\mathbf k)=0$ if $\mathbf k\notin\mathbf f(\Omega)$.

{\bf Computation of $t(\mathbf k)$.}
We are now ready to describe how to compute $t(\mathbf k)$.
Since $t(\mathbf k)=0$ for $\mathbf k\notin\mathcal B$, it suffices to compute $t(\mathbf k)$ for all $\mathbf k\in\mathcal B$.
Similar to the construction of $\mathbf f$, we recursively compute
$$t_i(\mathbf k):=\sum_{\mathbf x\in \mathbf{f}^{-1}(\mathbf k)\cap \mathcal{S}_i}\exp\big(\inp{\boldsymbol\theta}{\mathbf x}\big),$$
i.e., $t_n(\mathbf k)=t(\mathbf k)$.
{The recursive computation of $t_i(\mathbf k)$ is based on the following claim.
\begin{claim}\label{claim:computet-1}
$t_i(\mathbf k)=t_{i-1}(\mathbf k)+\exp(2\theta_i)\cdot t_{i-1}(\mathbf k-[\widehat{u}_{ji}]_{j=1}^r)$.
\end{claim}
The proof of Claim \ref{claim:computet-1} is presented in the supplementary material.
The above claim implies that once 
$t_{i-1}(\mathbf k)$ for $\mathbf k\in\mathcal B$ is obtained, $t_i(\mathbf k)$ can be efficiently computed using $t_{i-1}(\mathbf k)$.
Here, we consider $t_{i-1}(\mathbf k)=0$ for $\mathbf k\notin\mathcal B$.
Initially, one can find $t_0(\mathbf k)$ as follows:
$$t_0(\mathbf k)=\begin{cases}&\exp\left(-\sum_{i=1}^n\theta_i\right)~~\text{if}~~\mathbf k=\mathbf f\big((-1,\dots,-1)\big)\\
&\qquad\qquad 0\qquad\quad\text{otherwise}\end{cases}$$
where $\mathbf f\big((-1,\dots,-1)\big)$ is defined in \eqref{eq:approxf-3}.}

\subsection{Provable Guarantee} 
\begin{algorithm}\caption{Spectral inference for low-rank GMs}\label{alg:rankrz}
\begin{algorithmic}[1]
\STATE {\bf Input:} 
$\boldsymbol\theta,\lambda_1,\dots,\lambda_r,\mathbf{v}_1,\dots,\mathbf{v}_r,c$
\STATE {\bf Output:} $\widehat Z$
\STATE $\mathbf{u}_j\leftarrow\sqrt{|\lambda_j|}\mathbf{v}_j$ for all $j\in\{1,\dots,r\}$
\STATE $b_j\leftarrow\big\lceil\|\mathbf{u}_j\|_1/c+(n+1)/2\big\rceil$ for all $j\in\{1,\dots,r\}$
\STATE $\mathcal B\leftarrow \prod_{j=1}^r\{-b_j,\dots,b_j\}$
\STATE $t(\mathbf k)\leftarrow 0$ for all $\mathbf k\in\mathcal B$
\STATE $\boldsymbol{\ell}\leftarrow[\ell_j]_{j=1}^r:\ell_j\leftarrow\arg\min_{\ell_j\in\mathbb{Z}} \left|c\cdot\ell_j+\sum_{i=1}^nu_{ji}\right|$
\STATE $t(\boldsymbol \ell)\leftarrow \exp(-\sum_i\theta_i)$
\FOR{$1\le i\le n$}
\STATE $\widehat u_{ji}\leftarrow\arg\min_{\widehat u_{ji}\in\mathbb{Z}} |c\cdot\widehat u_{ji}-2u_{ji}|$
\STATE $t^\prime(\mathbf k)\leftarrow \exp(2\theta_i)t(\mathbf k-[\widehat u_{ji}]_{j=1}^r)$ for all $\mathbf k\in\mathcal  B$\footnotemark
\STATE $t(\mathbf k)\leftarrow t(\mathbf k)+t^\prime(\mathbf k)$ for all $\mathbf k\in\mathcal  B$
\ENDFOR
\STATE $\widehat Z\leftarrow\sum_{\mathbf k\in\mathcal B}t(\mathbf k)\exp\left(\sum_{j=1}^r\text{sign}(\lambda_j)(c\cdot k_j)^2\right)$
\end{algorithmic}
\end{algorithm}
\footnotetext{
Choose $t(\mathbf k)=0$ for $\mathbf k\notin\mathcal B$.}
The succinct description of the
proposed approximate inference algorithm described in Section \ref{sec:overspec} and \ref{sec:choicef}
is given in Algorithm \ref{alg:rankrz}.
We further prove the following theoretical guarantee of the algorithm.

\begin{theorem}\label{thm:rankrz}
Algorithm \ref{alg:rankrz} outputs $\widehat Z$ such that
\begin{align*}
&\left|\log\frac{Z}{\widehat Z}\right|
\le\frac14r{c^2(n+1)^2}+c\sqrt{n}(n+1)\sum_{j=1}^r\sqrt{|\lambda_j|},
\end{align*}
in $O\big(n2^r\prod_{j=1}^r(\sqrt{|\lambda_j|n}/c+n/2+1)\big)$ time.
\end{theorem}
The proof of 
Theorem \ref{thm:rankrz} is presented in the supplementary material.
As expected, a smaller quantization interval $c$ provides a smaller error bound, but a higher complexity
(and vice versa).
From Theorem \ref{thm:rankrz}, given $\varepsilon \in (0,1/2)$, 
one can check that Algorithm \ref{alg:rankrz} guarantees
$$(1-\varepsilon) {Z}\leq \widehat{Z} \leq (1+\varepsilon) {Z},$$
if we choose $$c=\min\left(\sqrt{\frac{\varepsilon}{r}}\frac1{n+1},\frac\varepsilon{4(\sum_j\sqrt{|\lambda_j|})\sqrt{n}(n+1)}\right).$$
Under the choice of $c$, the algorithm complexity becomes $O\big((\frac9\varepsilon r\max(\lambda_{\max},1))^rn^{2r+1}\big)$ where $\lambda_{\max}=\max_j|\lambda_j|$.
Therefore, if the rank and parameters of GM are bounded, i.e., $r, A_{ij}=O(1)$ for all $i,j$,
Algorithm \ref{alg:rankrz} is a fully polynomial-time approximation scheme (FPTAS) for approximating $Z$.

Finally, we remark that  
the following simple trick allows a FPTAS for approximating the partition function of
a richer class of GMs: for any diagonal matrix $D$, one can check
\begin{equation}\label{eq:zdiag}
Z=Z(\boldsymbol\theta, A) = \exp\big(-\text{Tr}(D)\big)\cdot Z(\boldsymbol\theta, A+D).
\end{equation}
Namely, if there exists a diagonal matrix $D$ such that the rank of $A+D$ is $O(1)$ 
(possibly, $A$ is not of low-rank though),
then one can run Algorithm \ref{alg:rankrz} to approximate $Z(\boldsymbol\theta, A+D)$ and
use it to derive $Z(\boldsymbol\theta, A)$ from \eqref{eq:zdiag}.

\section{Spectral Inference for High-Rank GMs}
In the previous section, we introduced a FPTAS algorithm for approximating the partition function for the special class of low-rank GMs.
However, for general (high-rank) GMs, Algorithm \ref{alg:rankrz} is intractable to run 
as its complexity grows exponentially with respect to the rank.
In this section, we address the issue by proposing a new efficient partition function approximation algorithm for general GMs of arbitrary rank.
The proposed algorithm utilizes Algorithm \ref{alg:rankrz} as a subroutine.
Our main idea is to decompose the partition  function of GM into a product of that of rank-1 GMs using the mean-field approximation, and then handle each rank-1 GM via Algorithm \ref{alg:rankrz}.

Throughout this section, we assume GMs with $\boldsymbol\theta=0$.
Such a restriction does not harm the generality of our method 
due to the following:
\begin{align*}
Z&=\sum_{\mathbf x\in\Omega=\{-1,1\}^n}\exp\left(\inp{\boldsymbol\theta}{\mathbf x}+\mathbf{x}^TA\mathbf x\right)\\
&=\frac12\sum_{\mathbf{x}^\prime\in\{-1,1\}^{n+1}}\exp\left((\mathbf{x}^\prime)^TA^\prime \mathbf{x}^\prime\right)=\frac12Z^\prime
\end{align*}
where  $A^\prime=\begin{bmatrix}
A & \frac12\boldsymbol\theta\\
\frac12\boldsymbol\theta^T & 0
\end{bmatrix}$ and $Z^\prime=\sum_{\mathbf{x}^\prime}\exp\left((\mathbf{x}^\prime)^TA^\prime \mathbf{x}^\prime\right)$ is the partition function of a GM with $A^\prime$.
Namely, computing the partition function of any GM is easily reducible to computing 
that of an alternative GM with $\boldsymbol\theta=0$.

\subsection{Overall Approach: From High-Rank to Low-Rank}
To handle high-rank GMs, we first reformulate the partition function $Z$ by substituting the summation over $\mathbf x$ with the expectation over $\mathbf x$ drawn from the uniform distribution $U_\Omega$ over $\Omega$:
\begin{align}
Z&=\sum_{\mathbf x\in\Omega}\exp\left(\mathbf{x}^TA\mathbf x\right)=2^n\mathbb{E}_{\mathbf x\sim U_\Omega}\left[\exp\left(\mathbf{x}^TA\mathbf x\right)\right].\label{eq:meanfieldz}
\end{align}
Then, for approximating the above expectation, we consider the following mean-field approximation via some fully factorized distribution $q(\mathbf y)=\prod_{j=1}^n q_j(y_j)$, where $y_j=\inp{\mathbf v_j}{\mathbf x}$, $\mathbf y=[y_j]_{j=1}^n$:
\begin{align}
&\mathbb{E}_{\mathbf x\sim U_\Omega}\left[\exp\left(\mathbf{x}^TA\mathbf x\right)\right]
=\mathbb{E}_{\mathbf y\sim P_{\mathcal Y}}\left[\exp\left(\sum_{j=1}^n\lambda_jy_j^2\right)\right]\notag\\
&\approx\mathbb{E}_{\mathbf y\sim q}\left[\exp\left(\sum_{j=1}^n\lambda_jy_j^2\right)\right]
=\prod_{j=1}^n\mathbb{E}_{y_j\sim q_j}\left[\exp\left(\lambda_jy_j^2\right)\right],\label{eq:approx0}
\end{align}
where 
$P_{\mathcal Y}(\mathbf y):=\sum_{\mathbf x\in\Omega\,:\,y_j=\inp{\mathbf v_j}{\mathbf x},~\forall j}U_\Omega(\mathbf x)$ for $\mathbf y\in \mathcal Y$ and $\mathcal Y :=\big\{\mathbf y=[y_j=\inp{\mathbf v_j}{\mathbf x}]_{j=1}^n:\mathbf x\in\Omega\big\}$.
Now, we prove the following claim that 
the choice of $q_j(y_j)=P_{\mathcal Y}(y_j)$ (the marginal probability of the joint distribution $P_{\mathcal Y}$) is optimal for the mean-field approximation in \eqref{eq:approx0},
with respect to
the Kullback-Leibler (KL) divergence. The proof of Claim \ref{claim:kl} is presented in the supplementary material.
\begin{claim}\label{claim:kl}
$\text{KL}\big(P_{\mathcal Y}(\mathbf y)||\prod_{j=1}^n q_j(y_j)\big)$ is minimized when $q_j(y_j)=P_{\mathcal Y}(y_j)$ for all $j$.
\end{claim}
In summary, under the choice of $q_j(y_j)=P_{\mathcal Y}(y_j)$, we use the following approximation for $Z$
from \eqref{eq:meanfieldz} and \eqref{eq:approx0}:
\begin{align}
Z&=2^n\mathbb{E}_{\mathbf x\sim U_\Omega}\left[\exp\left(\mathbf x^TA\mathbf x\right)\right]\notag\\
&\approx2^n\prod_{j=1}^n\mathbb{E}_{y_j\sim q_j}\left[\exp\left(\lambda_jy_j^2\right)\right]\notag\\
&=2^n\prod_{j=1}^n
\mathbb{E}_{\mathbf x\sim U_\Omega}\left[\exp\left(\lambda_j\inp{\mathbf v_j}{\mathbf x}^2\right)\right],\label{eq:approx1}
\end{align}
where it is easy to check that
$2^n\mathbb{E}_{\mathbf x\sim U_\Omega}\big[\exp\big(\lambda_j\inp{\mathbf v_j}{\mathbf x}^2\big)\big]$ 
is equivalent to the partition function of a rank-1 GM induced by $\lambda_j, \mathbf v_j$ and
can be efficiently approximated using Algorithm \ref{alg:rankrz}.
We further remark that the mean-field approximation quality in \eqref{eq:approx1} is
expected to be better if 
variables $y_j=\inp{\mathbf v_j}{\mathbf x}$ for all $j$ are closer to independence. 
Hence, it is quite a reasonable approximation
since for $i\ne j$, $\inp{\mathbf v_i}{\mathbf x}$, $\inp{\mathbf v_j}{\mathbf x}$ are pairwise uncorrelated, i.e., $\mathbb{E}_{\mathbf x\sim U_\Omega}[\inp{\mathbf v_i}{\mathbf x}\inp{\mathbf v_j}{\mathbf x}]=0$, due to the orthogonality of eigenvectors $\mathbf v_i,\mathbf v_j$. 

{We remark that our mean-field approximation \eqref{eq:approx1} is different from the traditional one \cite{parisi1988statistical}.
The latter addresses to find a mean-field distribution of $x_i$'s minimizing the KL divergence with the original distribution $\mathbb{P}(\mathbf x)$, while our approach minimizes the KL divergence between $q(y_j)$ and $P_{\mathcal Y}(\mathbf y)$, i.e., after spectral processing.}

\subsection{Improving \eqref{eq:approx1} via Controlling the Diagonal of $A$}
It is instructive to remind that varying the diagonal of $A$ only changes the partition function by a constant multiplicative factor, as in \eqref{eq:zdiag}. In order to fully utilize this, we address to optimize the diagonal of $A$ to improve our mean-field approximation.
To this end,
we build the following mean-field approximation by introducing the additional freedom of choosing a diagonal matrix $D$:
\begin{align}
&\mathbb{E}_{\mathbf x\sim U_\Omega}\left[\exp\left(\mathbf x^TA\mathbf x\right)\right]\notag\\
&=\mathbb{E}_{\mathbf x\sim U_\Omega}\left[\exp\left(\mathbf x^T(A+D)\mathbf x-\text{Tr}(D)\right)\right]\notag\\
&=\mathbb{E}_{\mathbf y^D\sim  P_{\mathcal{Y}^D}}\left[\exp\left(\sum_{j=1}^n\lambda^D_j(y^D_j)^2-\text{Tr}(D)\right)\right]\notag\\
&\approx\mathbb{E}_{\mathbf y^D\sim q^D}\left[\exp\left(\sum_{j=1}^n\lambda^D_j(y^D_j)^2-\text{Tr}(D)\right)\right]\label{eq:optd1}
\end{align}
where 
$\lambda^D_j,\mathbf y^D,q^D,\mathcal{Y}^D,P_{\mathcal{Y}^D}$ are those for $A+D$ (analogous to $\lambda_j,\mathbf y,q,\mathcal{Y},P_{\mathcal{Y}}$ of $A$).
Since it is intractable to find the optimal selection for $D$ by directly minimizing the approximation gap of \eqref{eq:optd1} (as computing the true expectations is intractable), we propose to set the free parameter $D$ by solving the following semi-definite programming (SDP): 
\begin{equation}\label{eq:sdp}
\max_{D}\text{Tr}(D)\qquad\text{subject to}\qquad A+D\preceq0.
\end{equation}
The intuition behind solving \eqref{eq:sdp} is provided in Section \ref{sec:sdpintuition}. We also provide its empirical justification through experimental studies in Section \ref{sec:sdp}.
We remark that the SDP \eqref{eq:sdp} is equivalent to (the dual of) the popular semi-definite relaxation of the max-cut problem \cite{goemans1995improved} and the maximum eigenvalue minimization problem \cite{delorme1993laplacian}.
For the complexity of solving \eqref{eq:sdp}, the interior point method \cite{alizadeh1995interior,helmberg1996interior} has $O(n^{3.5}\log(1/\varepsilon))$ running time and the first order method \cite{nesterov2007smoothing} has $O(n^3\sqrt{\log n}/\varepsilon)$ running time 
where $\varepsilon>0$ denotes the target precision to the optimum.\footnote{We also refer Section 3 of \cite{waldspurger2015phase} and Section 4 of \cite{goemans1995improved} for more details.}

From \eqref{eq:meanfieldz}, \eqref{eq:optd1} and \eqref{eq:sdp}, our final approximation becomes
\begin{align}
Z&=2^n\mathbb{E}_{\mathbf x\sim U_\Omega}\left[\exp\left(\mathbf x^TA\mathbf x\right)\right]\notag\\
&\approx 2^n\mathbb{E}_{\mathbf y^D\sim q^D}\left[\exp\left(\sum_{j=1}^n\lambda_j(y_j^D)^2-\text{Tr}(D)\right)\right]\notag\\
&=2^n\exp\left(-\text{Tr}(D)\right)\prod_{j=1}^{n}\mathbb{E}_{y_j^D\sim q_j^D}\left[\exp\left(\lambda_j(y_j^D)^2\right)\right]\notag\\
&=2^n\exp\left(-\text{Tr}(D)\right)\prod_{j=1}^{n}\mathbb{E}_{\mathbf x\sim U_\Omega}\left[\exp\left(\lambda_j^D\inp{\mathbf v_j^D}{\mathbf x}^2\right)\right]\notag
\end{align}
where $D$ is a solution of \eqref{eq:sdp} and $\mathbf v^D_j$ is an eigenvector of $A+D$ corresponding to $\lambda_j^D$. It is trivial that the above approximation with $D=0$
reduces to \eqref{eq:approx1}.
Finally, we formally state the proposed algorithm in Algorithm \ref{alg:approx}.

\begin{figure*}[ht]
\centering
    \begin{subfigure}{0.31\textwidth}
    \includegraphics[width=0.952\textwidth]{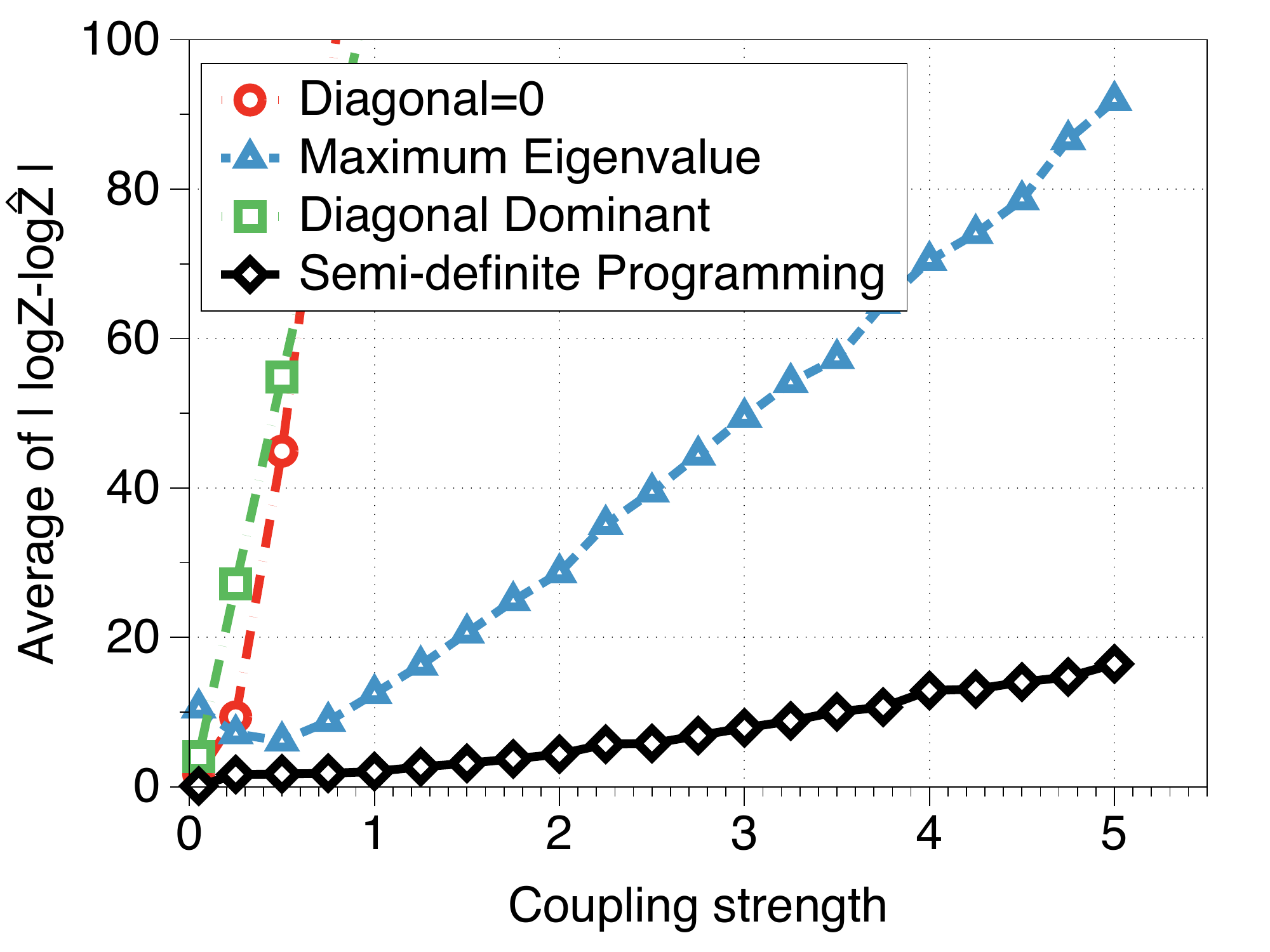}
    \caption{Approximation errors with different $D$}
    \label{fig:sdp}
    \end{subfigure}
    \begin{subfigure}{0.31\textwidth}
    \includegraphics[width=0.952\textwidth]{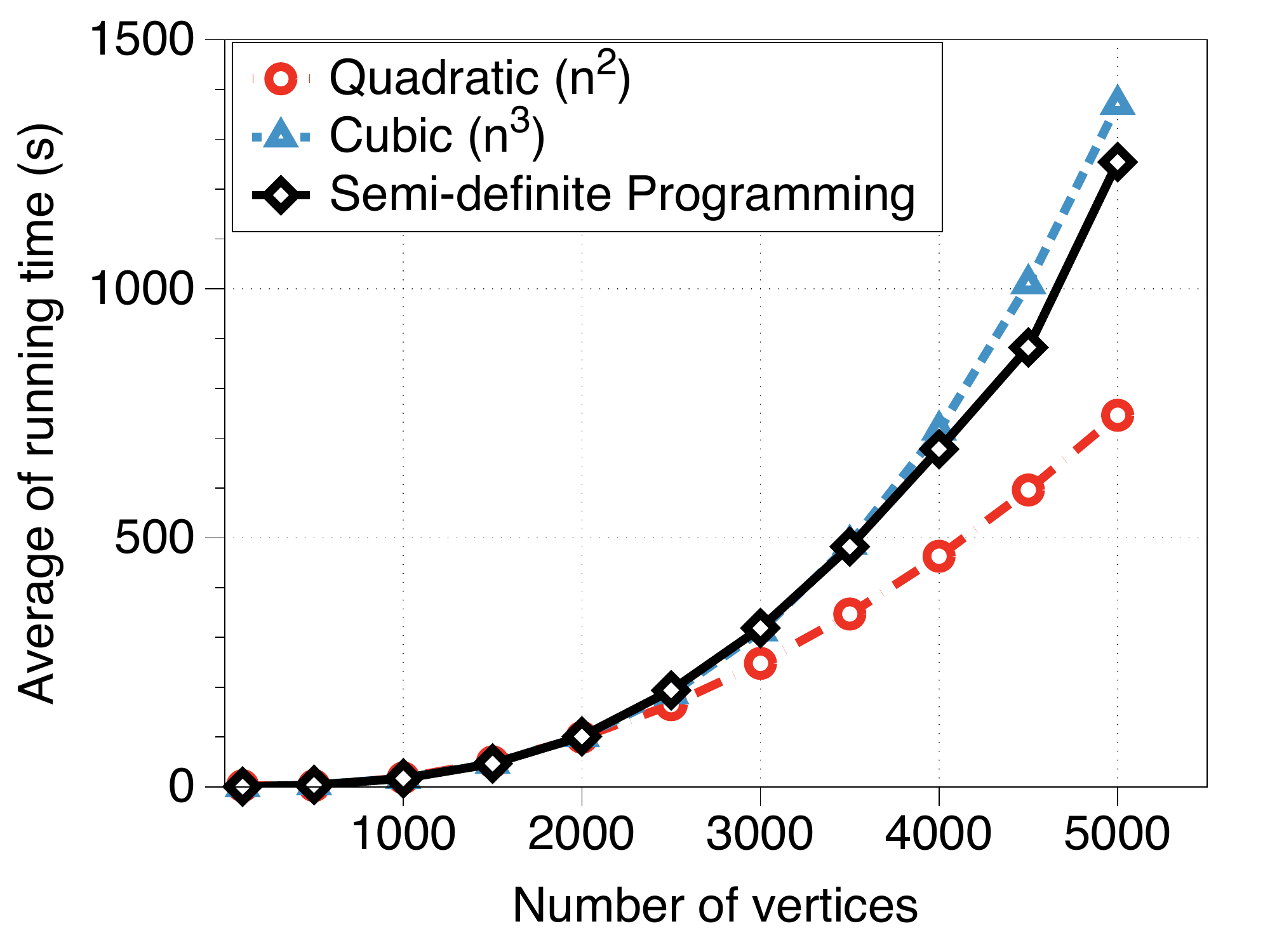}
    \caption{Running time for solving \eqref{eq:sdp}}
    \label{fig:sdptime}
    \end{subfigure}
    \begin{subfigure}{0.3\textwidth}
    \includegraphics[width=0.952\textwidth]{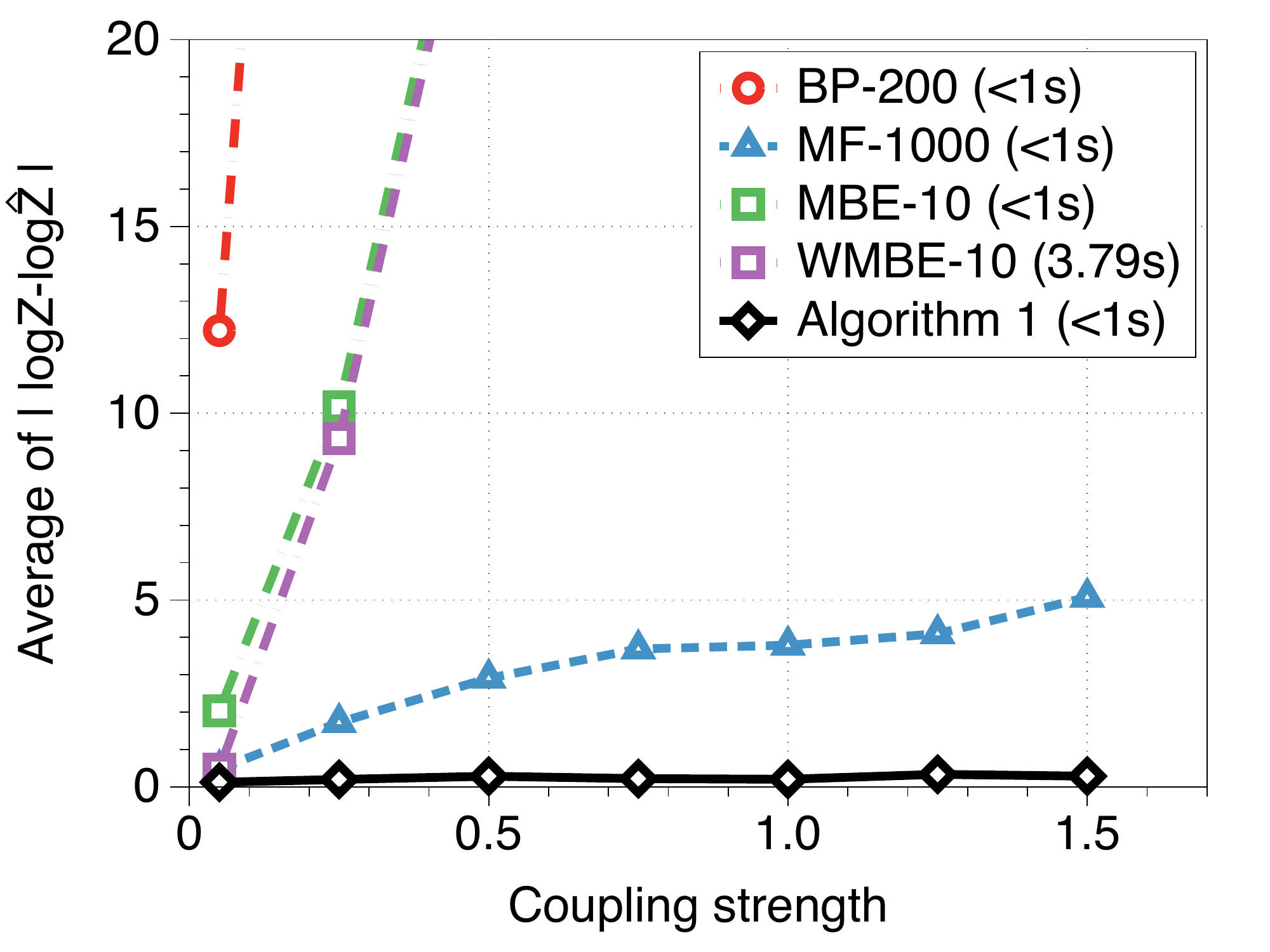}
    \caption{Approximation errors for rank-1 GMs}
    \label{fig:lowrank}
    \end{subfigure}
    \caption{Evaluation to measure (a) the effect and (b) running time 
    of the semi-definite programming \eqref{eq:sdp}. (c) Evaluation of Algorithm \ref{alg:rankrz} under rank-1 GMs.
    In (b), quadratic and cubic denote polynomial regression curves fitted for 100-2000 vertice results.}
\vspace{-0.05in}
\end{figure*}
\subsection{Intuition for \eqref{eq:sdp}}\label{sec:sdpintuition} 
Now, we describe the intuition why we consider the semi-definite programming \eqref{eq:sdp}.
To this end, let us re-write the approximation error in \eqref{eq:optd1} as the following alternative view:
\begin{align*}
&\left|\log(Z) - \log (\widehat{Z}) \right| \cr
&=\left|\log \left(\frac{\mathbb{E}_{\mathbf y^D\sim P_{\mathcal{Y}^D}}\left[\exp\left(\sum_{j=1}^n\lambda^D_j(y^D_j)^2+\text{Tr}(D)\right)\right]}{\mathbb{E}_{\mathbf y^D\sim q^D}\left[\exp\left(\sum_{j=1}^n\lambda^D_j(y^D_j)^2+\text{Tr}(D)\right)\right]} \right) \right|\cr 
&=\left|\log \left(\frac{\mathbb{E}_{\mathbf y^D\sim P_{\mathcal{Y}^D}}\left[\exp\left(\sum_{j=1}^n\lambda^D_j(y^D_j)^2\right)\right]}{\mathbb{E}_{\mathbf y^D\sim q^D}\left[\exp\left(\sum_{j=1}^n\lambda^D_j(y^D_j)^2\right)\right]} \right) \right|
\end{align*}
where $\widehat Z$ denotes the approximated partition function.
One can easily check that the approximation error is $0$ when $\lambda^D_1=\dots = \lambda^D_n =0$. Thus, we can expect a very accurate estimation when all eigenvalues of $A+D$ are close to 0. 
{One can also observe that if there exists $\lambda_j^D>0$, then the error might be too huge as $\sup_{y\in\mathbb{R}^n}\sum_{j=1}^n\lambda_j^Dy_j^2=\infty$
and the supports of $P_{\mathcal{Y}^D}$ and $q^D$ are different.
Under the above intuitions, we suggest to
solve the following problem:
\begin{equation}\label{eq:maxeig}
\max_{D} \sum_{j=1}^n\lambda^D_j \qquad\text{subject to}\qquad \lambda^D_j \le 0,~~\text{for all}~j.
\end{equation}
The optimization \eqref{eq:maxeig} is equivalent to \eqref{eq:sdp} since $\text{Tr}(D)=\sum_{j=1}^n\lambda_j^D - \text{Tr}(A)$ and the condition $\lambda^D_j \le 0$ for all $j$ is equivalent to $A+D\preceq0$.}

\begin{algorithm}[t]
\caption{Spectral inference for high-rank GMs}\label{alg:approx}
\begin{algorithmic}[1]
\STATE {\bf Input:} $A,c$
\STATE {\bf Output:} $\widehat Z$
\STATE $D\leftarrow$ solution of the semi-definite programming \eqref{eq:sdp}
\FOR{$1\le j\le n$}
\STATE $\lambda_j^D\leftarrow$ $j$-th largest eigenvalue of $A+D$
\STATE $\mathbf v_j^D\leftarrow$ $j$-th largest eigenvector of $A+D$
\STATE $E_j\leftarrow2^{-n}\times\text{Algorithm \ref{alg:rankrz}}(\theta=0,\lambda_j^D,\mathbf v_j^D,c)$
\ENDFOR
\STATE $\widehat Z\leftarrow2^n\times \exp\left(-\text{Tr}(D)\right)\times \prod_{j=1}^n E_j$
\end{algorithmic}
\vspace{-0.03in}
\end{algorithm}
\vspace{-0.05in}
\section{Experimental Results}\label{sec:exp}
In this section, we evaluate our algorithms on diverse environments including both synthetic and UAI datasets to corroborate our theorem and claims.
\vspace{-0.03in}
\subsection{Setups}
To begin with, we describe our overall experimental settings.
We compare our algorithms against the standard inference schemes dominantly used in most applications:
belief propagation (BP) \cite{pearl1982reverend}, mean-field approximation (MF) \cite{parisi1988statistical}, mini-bucket elimination (MBE) \cite{dechter2003mini} and weighted mini-bucket elimination (WMBE) \cite{liu2011bounding}.
Since all baselines are iterative methods and have the trade-off between the computation cost and the performance, we choose 200 iterations for BP, 1000 iterations for MF and 10 ibound for MBE and WMBE, for fair comparisons. Below these are referred to as BP-200, MF-1000, MBE-10 and WMBE-10, respectively.
In the case of BP and MF, their performances are saturated with the above choice in most cases and there is no gain by running more iterations. On the other hand, one can improve the approximation quality of MBE and WMBE with a larger ibound. However, its complexity grows exponentially with respect to it.
We also report
the running times of algorithms in our implementation
using round brackets following their names, 
e.g., BP-200 (2s) means that 200 iterations of BP run in 2 seconds (on average) for tested GMs.

Throughout our all experiments, we fix $c=\sqrt{|\lambda_j|}/1000$ for Algorithm \ref{alg:rankrz} and Algorithm \ref{alg:approx} 
to bound its running time regardless of eigenvalues.
For solving the semi-definite programming (SDP) \eqref{eq:sdp},
we use CVX \cite{grant2008cvx} with SDPT3 solver \cite{toh1999sdpt3} using MATLAB.

For generating synthetic GMs to evaluate on,
we first choose the graph structure (it will be specified in each setting) and randomly sample $\theta_i\sim\text{Unif}[-1,1]$ on its vertices and $A_{ij}\sim\text{Unif}[-s,s]$ on its edges
where $\text{Unif}$ denotes the uniform distribution and $s$ indicates the strength of pairwise couplings.
For measuring the running time for all experiments, we run algorithms using a single thread of CPU.
To reduce experimental noise, we average 100 random GMs for each plot unless otherwise stated.
\begin{figure*}[t]
    \centering
    \begin{subfigure}{0.31\textwidth}
    \includegraphics[width=0.952\textwidth]{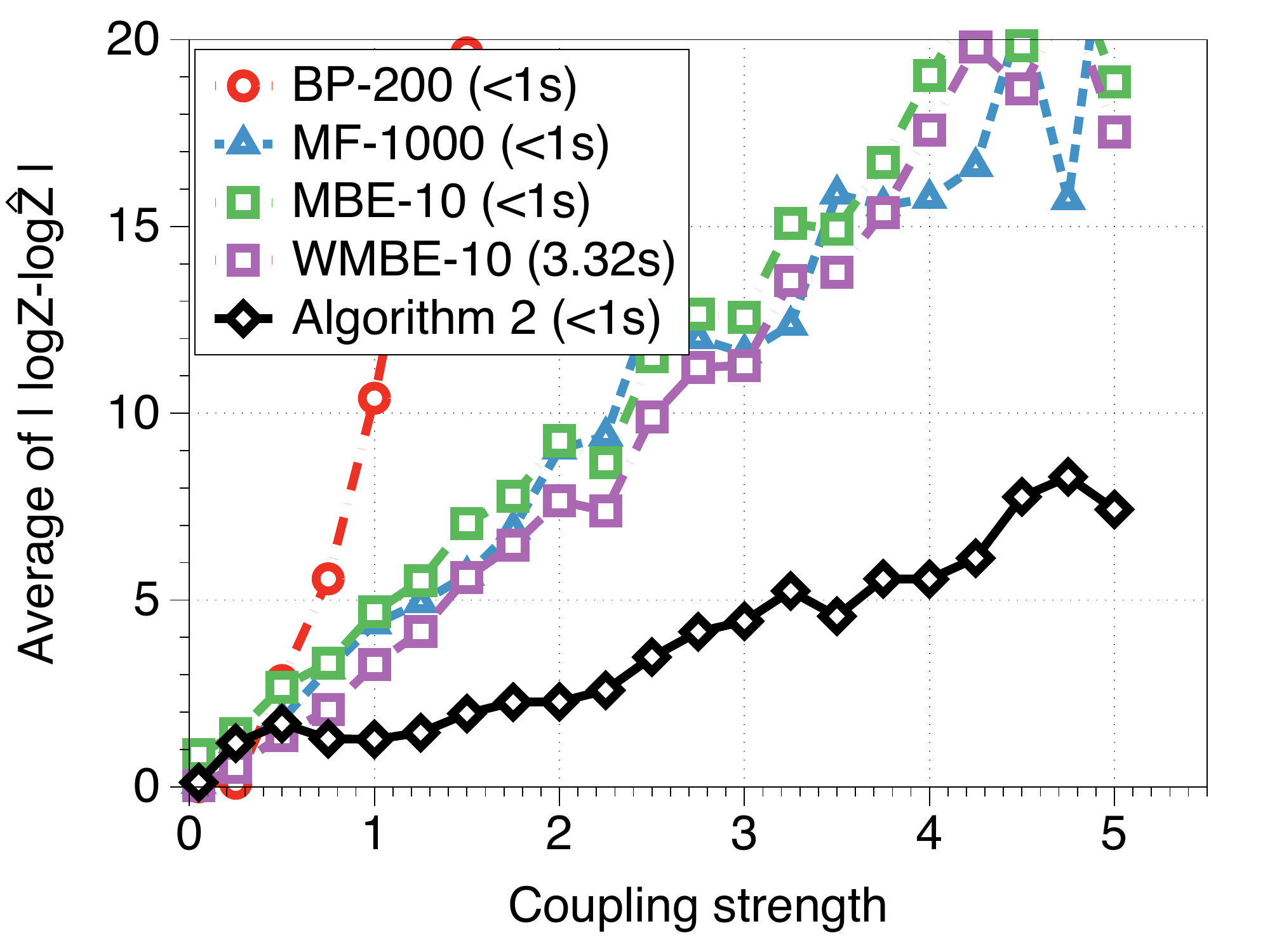}
    \caption{ER graph of 20 vertices: $p=0.5$}
    \label{fig:er05}
    \end{subfigure}
    \begin{subfigure}{0.31\textwidth}
    \includegraphics[width=0.952\textwidth]{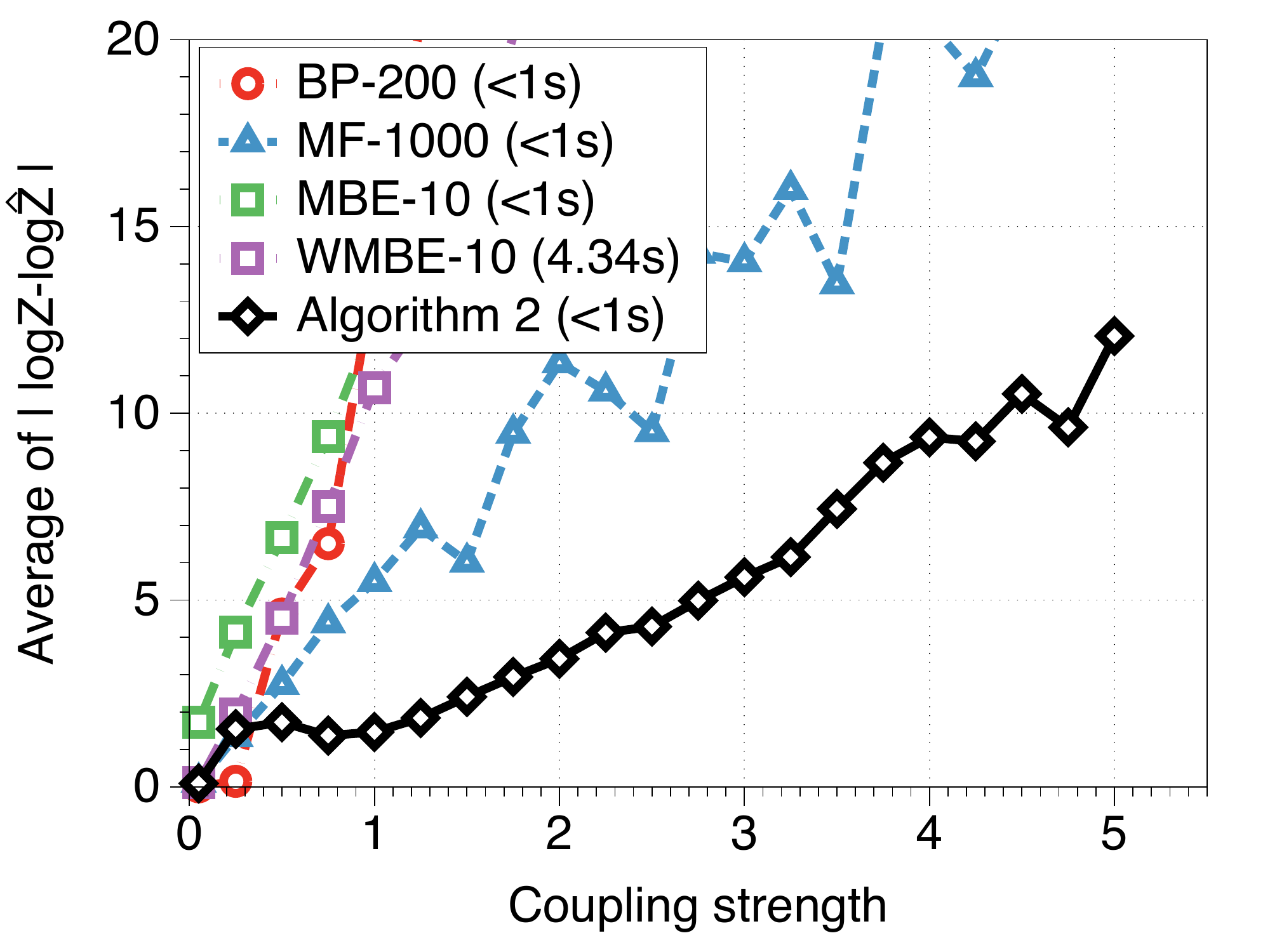}
    \caption{ER graph of 20 vertices: $p=0.7$}
    \end{subfigure}
    \begin{subfigure}{0.31\textwidth}
    \includegraphics[width=0.952\textwidth]{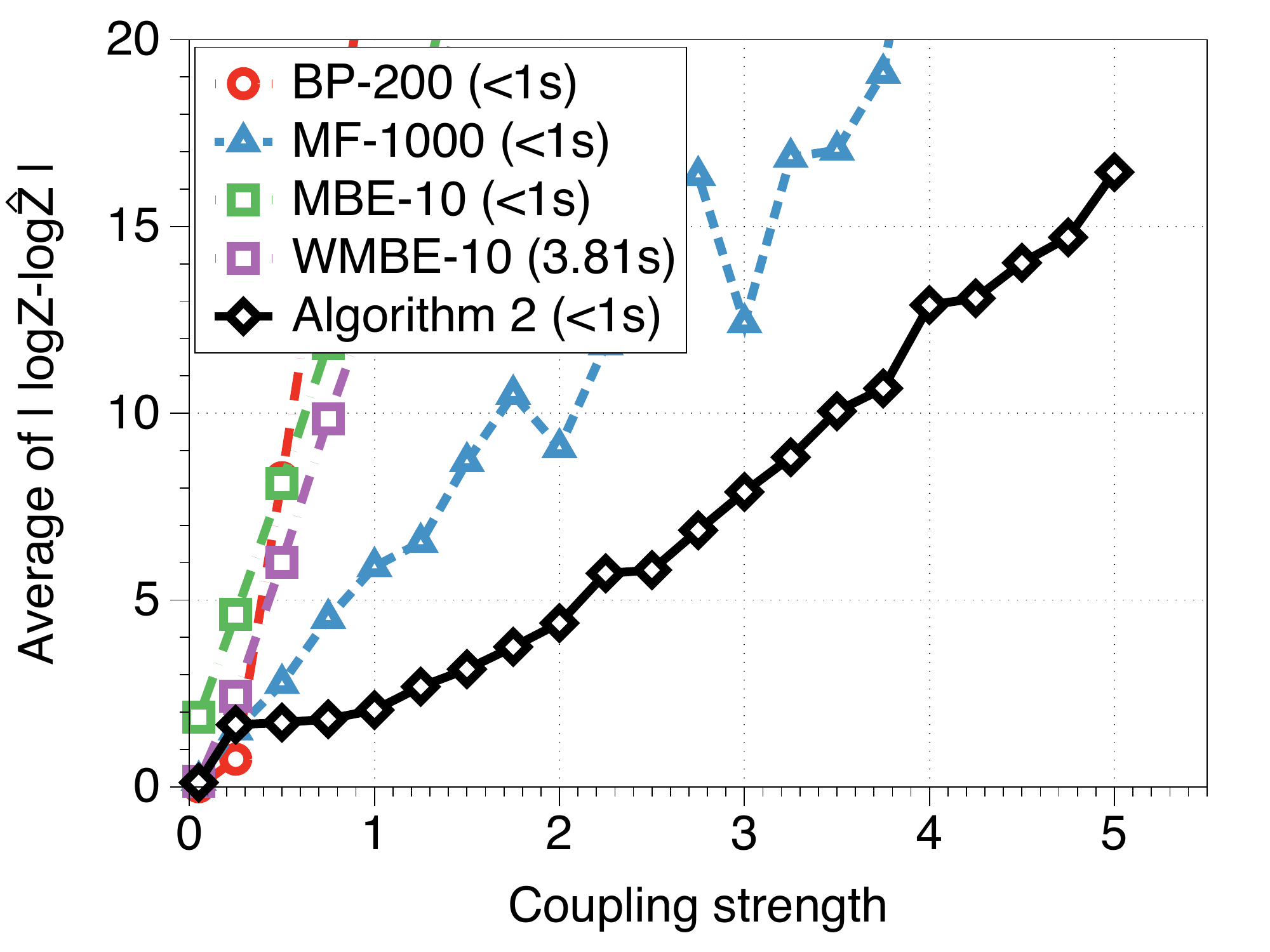}
    \caption{Complete graph of 20 vertices}
    \end{subfigure}
    \begin{subfigure}{0.31\textwidth}
    \includegraphics[width=0.952\textwidth]{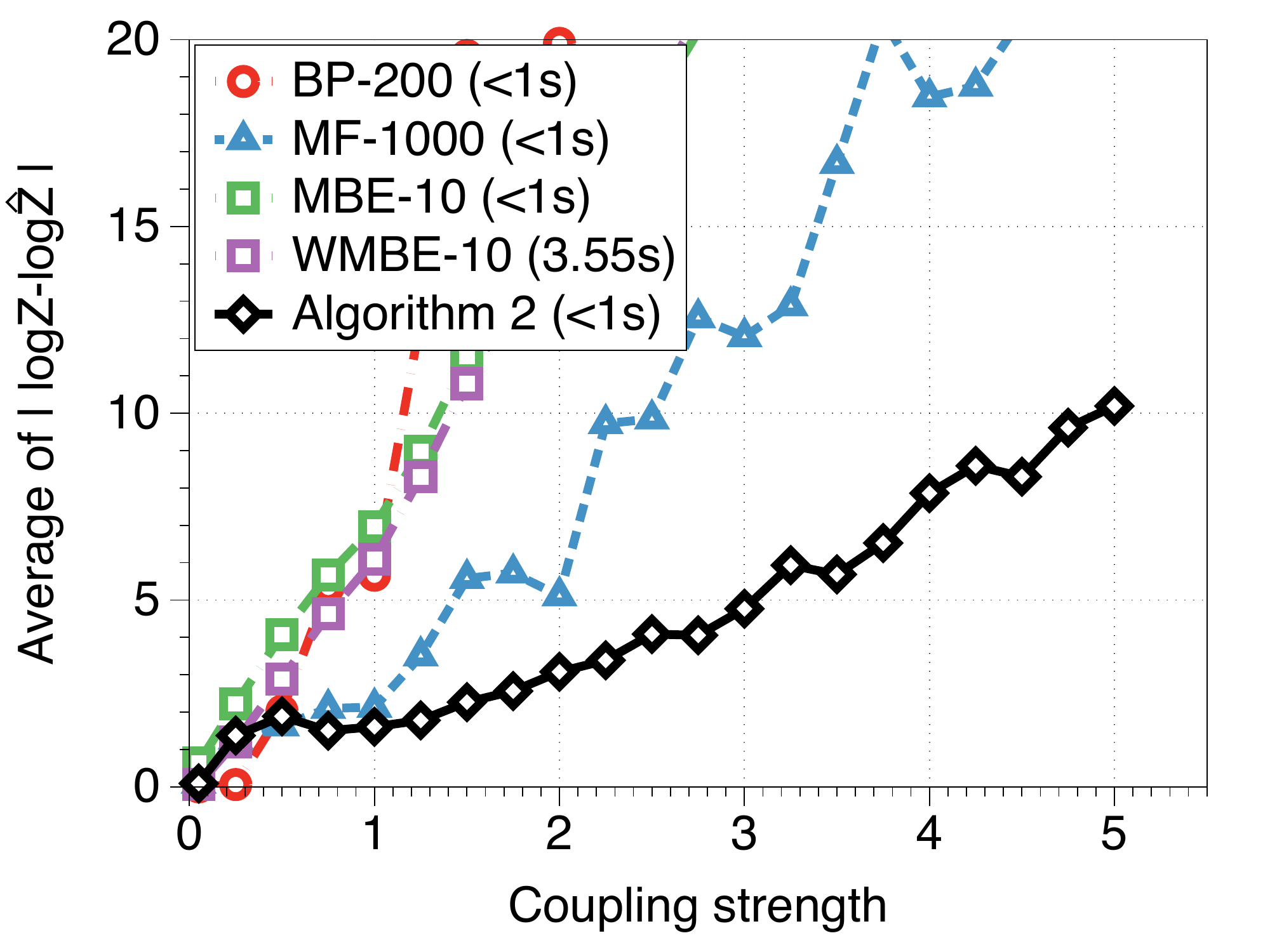}
    \caption{Complete bipartite graph of 20 vertices}
    \end{subfigure}
    \begin{subfigure}{0.31\textwidth}
    \includegraphics[width=0.952\textwidth]{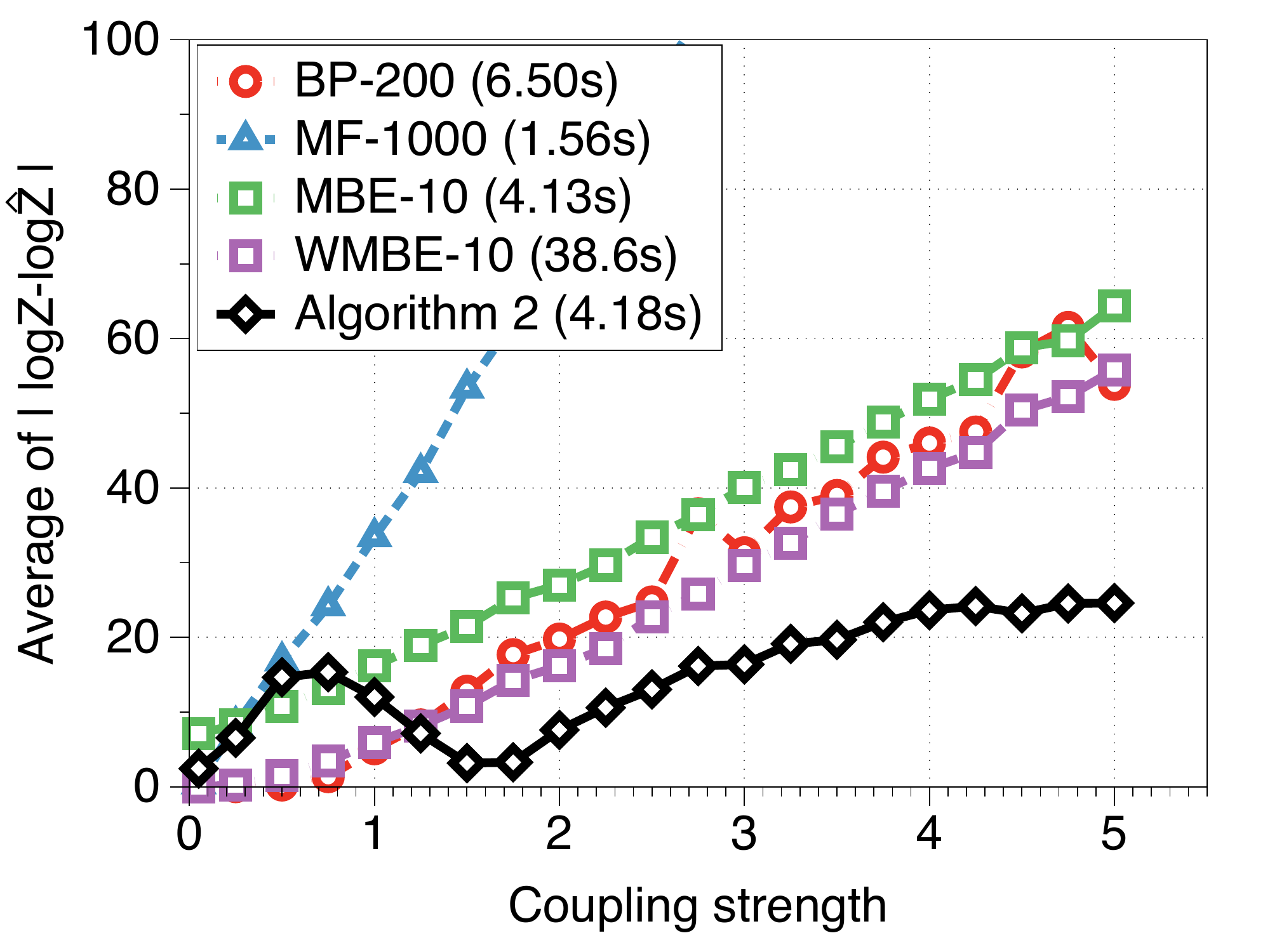}
    \caption{Grid graph of 225 vertices ($15\times15$)}
    \label{fig:grid}
    \end{subfigure}
    \begin{subfigure}{0.31\textwidth}
    \includegraphics[width=0.952\textwidth]{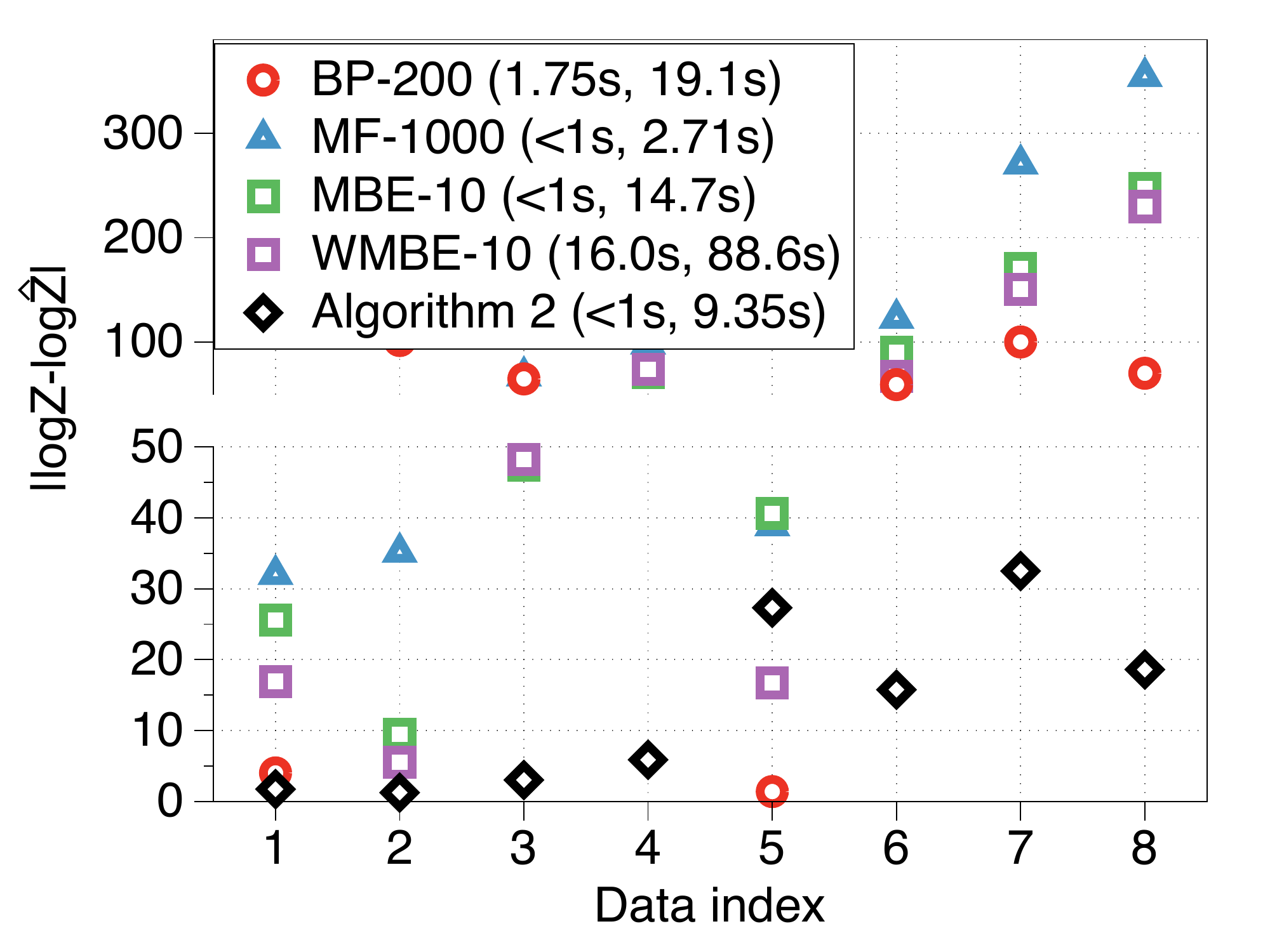}
    \caption{UAI grid GMs}
    \label{fig:uai}
    \end{subfigure}
    \caption{{Approximation errors for GMs on various graph structures and UAI datasets. 
    In (a)-(b), $p$ denotes the probability to choose each edge of ER graphs.
    In (f) for UAI grid GMs, indices 1-4 correspond to 
    GMs of 100 vertices ($10\times10$ grid)  and 5-8 correspond to GMs
    of 400 vertices ($20\times20$ grid). In (f), average running times of algorithms for indices 1-4 and indices 5-8 are noted in $(\cdot)$, respectively.}
    }
    \vspace{-0.1in}
    \label{fig:synthetic}
\end{figure*}
\subsection{Investigating the Semi-Definite Programming \eqref{eq:sdp}}\label{sec:sdp}
In this section, 
we investigate empirical effects and running time of the proposed SDP \eqref{eq:sdp}.

\noindent{\bf Effect of solving \eqref{eq:sdp}.}
We first investigate how \eqref{eq:sdp} helps the mean-field approximation \eqref{eq:optd1} used in Algorithm \ref{alg:approx} compared to other choices of diagonal matrix $D$.
In particular, we consider three other choices to compare.
The first choice is $D=0$ which does not change the diagonal of $A$.
The second choice is $D=-\max_j\lambda_j\times I$ which chooses entries of $D$ by the maximum eigenvalue of $A$ so that $A+D\preceq0$.
The last choice is $D_{ii}=-\sum_{j=1}^n|A_{ij}|$ which forces $A+D$ to be a diagonal dominant matrix, i.e., $A+D\preceq0$.
The second and third choices can be thought as feasible, yet non-optimal solutions of \eqref{eq:sdp}.
Figure \ref{fig:sdp} reports the experimental result for measuring the log partition error for GMs on complete graph having 20 vertices.
One can observe that solving \eqref{eq:sdp} is important
for the approximation performance of Algorithm \ref{alg:approx}.

\noindent{\bf Running time for solving \eqref{eq:sdp}.}
Now, we discuss about the empirical complexity of solving \eqref{eq:sdp}.
Our solver SDPT3 uses the primal-dual interior point method \cite{toh1999sdpt3} for solving \eqref{eq:sdp}.
To measure the running time of the solver, we generate random GMs on complete graphs  
by varying the number of vertices from $100$ to $5000$.
Figure \ref{fig:sdptime} illustrates the average running time of our solver  where each point is averaged over 10 random GMs.
We compare the running time of our solver with quadratic and cubic polynomials with respect to $n$.
One can observe that the empirical running time to solve
\eqref{eq:sdp}
is between $O(n^2)$ and $O(n^3)$, which is better than the theoretical bound of the interior point method $O(n^{3.5})$ \cite{helmberg1996interior}. 

\subsection{Evaluation of Algorithm \ref{alg:rankrz} under Low-Rank GMs}
We evaluate Algorithm \ref{alg:rankrz} under rank-1 GMs,
which is used as a subroutine of Algorithm \ref{alg:approx}.
We choose a random eigenvalue $\lambda\in\{-1,1\}$ and a random eigenvector $\mathbf v\in\text{Unif}\big(\{\mathbf v\in\mathbb{R}^n:\|\mathbf v\|_2=1\}\big)$ to generate rank-1 GMs by choosing $A=\lambda \mathbf v\mathbf v^T$ and $\theta_i\sim\text{Unif}[-1,1]$.
Given $\mathbf v$, we scale $\lambda$ to match the average value of $|A_{ij}|$ to be equal to some constant $s$ (coupling strength in Figure \ref{fig:lowrank}), i.e.,
$$\frac1{n(n+1)}\sum_{i=1}^n\sum_{j=1:i\ne j}^n|A_{ij}|=s.$$
We remark that
rank-1 GMs has the special property that if its eigenvalue $\lambda$ is positive (or negative), they are equivalent to ferromagnetic (or antiferromagnetic) models, i.e., $A_{ij}\ge0$ (or $A_{ij}\le0$) for $i\ne j$, respectively.
Figure \ref{fig:lowrank} reports the algorithm performances under rank-1 GMs.
As expected from our theoretical results (Theorem \ref{thm:rankrz}),
our algorithm is nearly exact, while other algorithms fail.
In particular, BP,  MBE and WMBE output very poor approximation since they usually fail in antiferromagnetic cases, i.e., negative eigenvalue.
The superior performance of Algorithm \ref{alg:rankrz} under rank-1 GMs implies that
the approximation error of Algorithm \ref{alg:approx} would mainly come from
the mean-field approximation \eqref{eq:optd1}.

\subsection{Evaluation of Algorithm \ref{alg:approx} under High-Rank GMs}\label{sec:exphighrank}
We now evaluate the empirical performance of Algorithm \ref{alg:approx} under synthetic high-rank GMs and UAI datasets \cite{gogate2014uai}.
In all cases, we have checked through simulations that BP and MF do not have better accuracy than BP-200 and MF-1000, respectively, even if we run the algorithms with much longer iterations.

To generate synthetic GMs, we consider Erd\H{o}s-R\'enyi (ER) random graphs, complete bipartite graphs, complete graphs, and grid graphs.
The experimental results are reported in Figure \ref{fig:er05}-\ref{fig:grid}.
In all cases, one can observe that our algorithm significantly
outperforms others in the high coupling region, i.e., the low-temperature regime.
It is known that MF outputs better approximations than others as the underlying graph structure becomes dense, e.g., complete graph \cite{ellis1978statistics}, however, our algorithm remarkably 
performs better than MF even in such cases. 
In particular, MF and BP exhibit high variance on their approximation errors in high coupling regions, while ours does not.

We also evaluate our algorithms with GMs on grid graphs in a dataset for UAI 2014 inference competition.
It provides 8 GMs on grid graphs,
where 4 of them are of 100 vertices ($10\times10$) and the other 4 
are of 400 vertices ($20\times20$).
Figure \ref{fig:uai} reports the approximation error and the running time of each algorithm.
In the experimental results, our algorithm consistently has small errors,
while other algorithms often fail badly.

\begin{figure}[ht]
\centering
\includegraphics[width=0.31\textwidth]{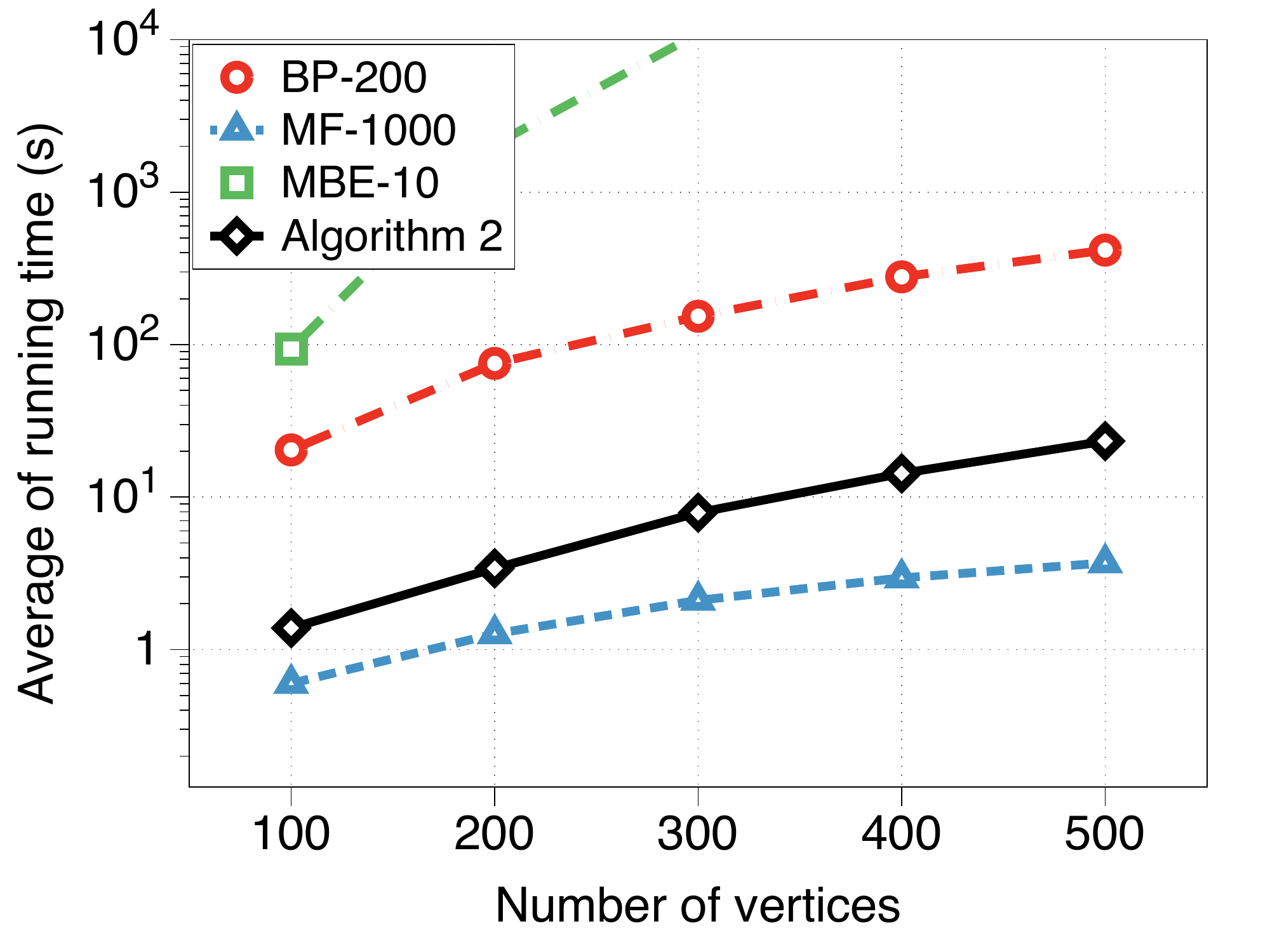}
    \caption{Running time under varying the size of GMs on complete graphs.}
    \label{fig:time}
\end{figure}

Finally, we compare the running times of algorithms under GMs on complete graphs of 100-500 vertices, which 
are reported in Figure \ref{fig:time}. Here, we do not report WMBE since it is slower than MBE.
One can observe that Algorithm \ref{alg:approx} scales as well as BP,
while MBE does not.
MF is the fastest, but 
it is worst in approximation quality under grid and UAI GMs, as reported in earlier experimental results.

\section{Conclusion}
In this paper, we provide a completely new angle to design
approximate inference algorithms for graphical models.
The proposed algorithms scale well for large scale models as like
prior iterative message-passing schemes, and
outperforms them in approximation quality, in particular, significantly for hard instances.
For the future work, we plan to extend our spectral approach to estimating the marginal distributions or/and related inference in higher-order or continuous models.

\section*{Acknowledgement}
This work was supported by IITP grant funded by the Korea government (MSIT) (No.2017-0-01779, XAI).
We would like to acknowledge Sungsoo Ahn for helpful discussions and sharing codes.

\bibliography{reference}
\bibliographystyle{icml2019}
\appendix
\clearpage
\onecolumn

\begin{center}{\bf {\LARGE Supplementary Material:}}
\end{center}

\begin{center}{\bf {\Large Spectral Approximate Inference}}
\end{center}
\vspace{0.3in}
\section{Proof of Claim \ref{claim:b}}
We first prove $\mathbf{f}(\Omega)\subset\mathcal B$.
To this end we introduce the following inequalities for all $\mathbf{x}\in\{-1,1\}^n$:
\begin{align}\label{eq:pfclm1}
|\inp{\mathbf u_j}{\mathbf{x}}|\le \|\mathbf u_j\|_1,\qquad |\inp{\mathbf u_j}{\mathbf{x}}-c\cdot f_j(\mathbf{x})|\le c\cdot\frac{n+1}2
\end{align}
which directly leads us to $|c\cdot f_j(\mathbf{x})|\le \|\mathbf u_j\|_1+c\cdot(n+1)/2\le c\cdot b_j$, and therefore $\mathbf{f}(\Omega)\subset\mathcal B$.
Here, the first inequality of \eqref{eq:pfclm1} is trivial.
The second inequality of \eqref{eq:pfclm1} is from the fact that the error between $c\cdot f_j(\mathbf{x})$ and $\inp{\mathbf u_j}{\mathbf{x}}$
arises from a series of quantizations which is presented once in \eqref{eq:approxf-3} and at most $n$ times in \eqref{eq:approxf-2}.
Since the quantization error is at most $c/2$ for each quantization, the second inequality of \eqref{eq:pfclm1} holds.

Now we prove the bound of $|\mathcal B|$.
From the definition of $\mathcal B$ and $b_j$, one can easily observe that the following bound on $|\mathcal B|$ holds:
\begin{align*}
|\mathcal B|=\prod_{j=1}^r(2b_j+1)&= 2^r\prod_{j=1}^r\left(\frac1c\|\mathbf u_j\|_1+\frac{n}2+1\right)\\
&= 2^r\prod_{j=1}^r\left(\frac1c\sqrt{|\lambda_j|}\|\mathbf v_j\|_1+\frac{n}2+1\right)\\
 &\le2^r\prod_{j=1}^r\left(\frac1c\sqrt{|\lambda_j|n}+\frac{n}2+1\right),
\end{align*}
where the inequality is from $\|\mathbf v_j\|_1\le\sqrt{n}\|\mathbf v_j\|_2=\sqrt{n}$.

\section{Proof of Claim \ref{claim:computet-1}}
Claim \ref{claim:computet-1} holds since
\begin{align}
t_i(\mathbf k)&=t_{i-1}(\mathbf k)+\sum_{\mathbf x\in \mathbf f^{-1}(\mathbf k)\cap (\mathcal{S}_i\setminus \mathcal{S}_{i-1})}\exp\big(\inp{\boldsymbol\theta}{\mathbf x}\big)\notag\\
&=t_{i-1}(\mathbf k)+\sum_{\mathbf g_i(\mathbf x)\in \mathbf g_i\big( \mathbf f^{-1}(\mathbf k)\cap (\mathcal{S}_i\setminus \mathcal{S}_{i-1})\big)}\exp\big(\inp{\boldsymbol\theta}{\mathbf x}\big)\notag\\
&=t_{i-1}(\mathbf k)+\sum_{\mathbf x^\prime\in \mathbf f^{-1}(\mathbf k-[\widehat{u}_{ji}]_{j=1}^r)\cap  \mathcal{S}_{i-1}}\exp\big(\inp{\boldsymbol\theta}{\mathbf g_i^{-1}(\mathbf x^\prime)}\big)\notag\\
&=t_{i-1}(\mathbf k)+\sum_{\mathbf x^\prime\in \mathbf f^{-1}(\mathbf k-[\widehat{u}_{ji}]_{j=1}^r)\cap  \mathcal{S}_{i-1}}\exp\big(2\theta_i+\inp{\boldsymbol\theta}{\mathbf x^\prime}\big)\notag\\
&=t_{i-1}(\mathbf k)+\exp(2\theta_i)\cdot t_{i-1}(\mathbf k-[\widehat{u}_{ji}]_{j=1}^r).\label{eq:computet-3}
\end{align}
In the above, $\mathbf g_i:\mathcal{S}_i\setminus \mathcal{S}_{i-1}\rightarrow \mathcal{S}_{i-1}$ is a bijection defined by $\mathbf g_i(\mathbf x)=\mathbf x^\prime$ such that $x^\prime_\ell=x_\ell$ except for $\ell=i$.
The second equality of \eqref{eq:computet-3} is from
replacing the summation over $\mathbf f^{-1}(\mathbf k)\cap (\mathcal{S}_i\setminus \mathcal{S}_{i-1})$ by that over $\mathbf g_i\big(\mathbf f^{-1}(\mathbf k)\cap(\mathcal{S}_i\setminus \mathcal{S}_{i-1})\big)$.
The third equality of \eqref{eq:computet-3}
is based on \eqref{eq:approxf-2} which implies that for all $\mathbf x\in \mathcal{S}_i\setminus \mathcal{S}_{i-1}$, $\mathbf x^\prime=\mathbf g_i(x)$ satisfies \begin{equation}\label{eq:computet-2}
\mathbf f(\mathbf x)-[\widehat{u}_{ji}]_{j=1}^r=\mathbf f(\mathbf x^\prime).
\end{equation}
Hence, \eqref{eq:computet-2} leads us to 
\begin{align*}
\mathbf g_i\big(\mathbf f^{-1}(\mathbf k)\cap(\mathcal{S}_i\setminus \mathcal{S}_{i-1})\big)&=\mathbf g_i\big(\{\mathbf x\in \mathcal{S}_i\setminus \mathcal{S}_{i-1}:\mathbf f(\mathbf x)=\mathbf k\}\big)\\
&=\{\mathbf x^\prime\in  \mathcal{S}_{i-1}:\mathbf f(\mathbf x^\prime)=\mathbf k-[\widehat{u}_{ji}]_{j=1}^r\}\\
&=\mathbf f^{-1}(\mathbf k-[\widehat{u}_{ji}]_{j=1}^r)\cap \mathcal{S}_{i-1}
\end{align*}
and the third equality of \eqref{eq:computet-3} follows.
The fourth equality of \eqref{eq:computet-3} directly follows from the definition of $\mathbf g_i$ that $x^\prime_i=-1$ and $\big(\mathbf g_i^{-1}(\mathbf x^\prime)\big)_i=x_i=1$.

\section{Proof of Theorem \ref{thm:rankrz}}
We first prove the computational complexity of Algorithm \ref{alg:rankrz}. 
Since each $t(\mathbf k), t^\prime(\mathbf k)$ possesses a memory of $O(|\mathcal B|)$ and $|\mathcal B|\le2^r\prod_{j=1}^r(\sqrt{|\lambda_j|n}/c+n/2+1)$ from Claim \ref{claim:b}, the space complexity of Algorithm \ref{alg:rankrz} is $O\big(2^r\prod_{j=1}^r(\sqrt{|\lambda_j|n}/c+n/2+1)\big)$.
In addition, as the algorithm iterates $n$ times while each iteration accesses to $t(\mathbf k)$ and $t^\prime(\mathbf k)$, Algorithm \ref{alg:rankrz} has $O\big(n2^r\prod_{j=1}^r(\sqrt{|\lambda_j|n}/c+n/2+1)\big)$ computational complexity.

Now we provide the bound on the partition function approximation.
First, we refer the following error bound introduced in the proof of Claim \ref{claim:b}.
\begin{equation}\label{eq:rankrzpf1}
\left|\inp*{\mathbf u_j}{\mathbf x}-c\cdot f_j(\mathbf x)\right|\le c\cdot\frac{n+1}2.
\end{equation}
Using \eqref{eq:rankrzpf1}, we provide a bound for $|\inp*{\mathbf u_j}{\mathbf x}^2-(c\cdot f_j(\mathbf x))^2|$ as follows
\begin{align}
|\inp*{\mathbf u_j}{\mathbf x}^2-(c\cdot f_j(\mathbf x))^2|\notag&=|\inp*{\mathbf u_j}{\mathbf x}-c\cdot f_j(\mathbf x)||\inp*{\mathbf u_j}{\mathbf x}+c\cdot f_j(\mathbf x)|\notag\\
&\le c\cdot\frac{n+1}2\left(|2\inp{\mathbf u_j}{\mathbf x}|+c\cdot\frac{n+1}2\right)\notag\\
&\le \frac14{c^2(n+1)^2}+c\sqrt{|\lambda_j|n}(n+1)\label{eq:rankrzpf2}
\end{align}
where the first inequality is from \eqref{eq:rankrzpf1} and the second inequality is from 
$|\inp{\mathbf u_j}{\mathbf x}|\le\|\mathbf u_j\|_1\le\sqrt{|\lambda_j|n}$.
From \eqref{eq:rankrzpf2}, the error bound can be derived as
\begin{align}
\frac{Z}{\widehat{Z}}&=
\frac{\sum_{\mathbf x\in\Omega}\exp\left(\inp{\boldsymbol\theta}{\mathbf x}+\sum_{j=1}^r\text{sign}(\lambda_j)\inp*{\mathbf u_j}{\mathbf x}^2\right)}{\sum_{\mathbf x\in\Omega}\exp\left(\inp{\boldsymbol\theta}{\mathbf x}+\sum_{j=1}^r\text{sign}(\lambda_j)(c\cdot f_j(\mathbf x))^2\right)}\notag\\
&\le\max_{\mathbf x\in\Omega}\frac{\exp\left(\inp{\boldsymbol\theta}{\mathbf x}+\sum_{j=1}^r\text{sign}(\lambda_j)\inp*{\mathbf u_j}{\mathbf x}^2\right)}{\exp\left(\inp{\boldsymbol\theta}{\mathbf x}+\sum_{j=1}^r\text{sign}(\lambda_j)(c\cdot f_j(\mathbf x))^2\right)}\notag\\
&\le\max_{\mathbf x\in\Omega}\exp\left(\sum_{j=1}^r|\inp{\mathbf u_j}{\mathbf x}^2-(c\cdot f_j(x))^2|\right)\notag\\
&\le\exp\left(\frac14r{c^2(n+1)^2}+c\sqrt{n}(n+1)\sum_{j=1}^r\sqrt{|\lambda_j|}\right)\notag
\end{align}
where the last inequality follows from \eqref{eq:rankrzpf2}.
One can obtain a same bound for $\widehat{Z}/Z$ and this completes the proof of Theorem \ref{thm:rankrz}.

\section{Proof of Claim \ref{claim:kl}}
The result of Claim \ref{claim:kl} directly follows from the following inequality:
\begin{align*}
\text{KL}\Big(P_{\mathcal Y}(\mathbf y)\big|\big|\prod_{j=1}^rq_j(y_j)\Big)&=-\sum_{\mathbf y\in\mathcal Y}P_{\mathcal Y}(\mathbf y)\log\prod_{j=1}^rq_j(y_j)-H(P_{\mathcal Y}(\mathbf y))\\
&=-\sum_{\mathbf y\in\mathcal{Y}}P_{\mathcal Y}(\mathbf y)\sum_{j=1}^r\log q_j(y_j)-H(P_{\mathcal Y}(\mathbf y))\\
&=-\sum_{j=1}^r\sum_{y_j:\mathbf y\in\mathcal Y}P_{\mathcal Y}(y_j)\log q_j(y_j)-H(P_{\mathcal Y}(\mathbf y))\\
&\ge-\sum_{j=1}^r\sum_{y_j:\mathbf y\in\mathcal Y}P_{\mathcal Y}(y_j)\log P_{\mathcal Y}(y_j)-H(P_{\mathcal Y}(\mathbf y))
\end{align*}
where the last inequality follows from the source coding theorem \cite{shannon1948mathematical}.
\end{document}